\documentclass{article}

\usepackage{times}
\usepackage{epsf,graphicx} 
\usepackage{subfigure} 
\usepackage{amsfonts,amssymb,amsmath,amstext,bm}

\usepackage{natbib}

\usepackage{algorithm}
\usepackage{algorithmic}

 \usepackage[nohyperref,accepted]{icml2014}

\icmltitlerunning{Stopping Criteria in Contrastive Divergence: Alternatives to the Reconstruction Error}

\def\vec#1{\mbox{\boldmath$\displaystyle#1$}}
\def\e    {\displaystyle{e}}
\def\b    {\vec{b}}
\def\c    {\vec{c}}
\def\x    {\vec{x}}
\def\xx   {\stackrel{\thicksim}{\vec{x}}}
\def\y    {\vec{y}}
\def\h    {\vec{h}}
\def\W    {\vec{W}}

\begin{document} 

\twocolumn[
\icmltitle{Stopping Criteria in Contrastive Divergence: Alternatives to the Reconstruction Error}

\icmlauthor{David Buchaca Prats}{davidbuchaca@gmail.com}
\icmladdress{Departament de Llenguatges i Sistemes Inform\`atics,\\
            Universitat Polit\`ecnica de Catalunya,\\ Barcelona, Spain}
\icmlauthor{Enrique Romero Merino}{eromero@lsi.upc.edu}
\icmladdress{Departament de Llenguatges i Sistemes Inform\`atics,\\
            Universitat Polit\`ecnica de Catalunya,\\ Barcelona, Spain}
\icmlauthor{Ferran Mazzanti Castrillejo}{ferran.mazzanti@upc.edu}
\icmladdress{Departament de F\'{\i}sica i Enginyeria Nuclear,\\
            Universitat Polit\`ecnica de Catalunya,\\ Barcelona, Spain}
\icmlauthor{Jordi Delgado Pin}{jdelgado@lsi.upc.edu}
\icmladdress{Departament de Llenguatges i Sistemes Inform\`atics,\\
            Universitat Polit\`ecnica de Catalunya,\\ Barcelona, Spain}

\icmlkeywords{Restricted Boltzmann Machines, Contrastive Divergence, Machine Learning, ICLR}

\vskip 0.3in
]

\begin{abstract} 
  Restricted Boltzmann Machines (RBMs) are general unsupervised
  learning devices to ascertain generative models of data
  distributions. RBMs are often trained using the Contrastive
  Divergence learning algorithm (CD), an approximation to the gradient
  of the data log-likelihood. A simple reconstruction error is often used
  to decide whether the approximation provided by the CD algorithm is
  good enough, though several authors
  \cite{schulz-et-al-Convergence-Contrastive-Divergence-2010-NIPSw,
    fischer-igel-Divergence-Contrastive-Divergence-2010-ICANN} have
  raised doubts concerning the feasibility of this procedure. However,
  not many alternatives to the reconstruction error have been used in
  the literature. In this manuscript we investigate simple
  alternatives to the reconstruction error in order to detect as soon
  as possible the decrease in the log-likelihood during learning.
\end{abstract} 

\section{Introduction}
\label{introduction}

Learning algorithms for deep multi-layer neural networks have been
known for a long time \cite{RUMELHART-HINTON-WILLIAMS86}, though none
of them have been widely used to solve large scale real-world
problems. In 2006, Deep Belief Networks (DBNs)
\cite{hinton-et-al-DeepBeliefNetworks-2006-NC} came out as a real
breakthrough in this field, since the learning algorithms proposed
ended up being a feasible and practical method to train these
networks, with spectacular results
\cite{hinton-salakhutdinov-Auto-Encoders-2006-Science,
larochelle-et-al-Strategies-Train-Deep-Networks-2009-JMLR,
lee-at-al-Convolutional-Scalable-2009-ICML,le-at-al-google_unsupervised-2012-ICML}. DBNs
have Restricted Boltzmann Machines (RBMs)
\cite{smolensky-Restricted-Boltzmann-Machines-1986-PDP} as their
building blocks.

RBMs are topologically constrained Boltzmann Machines
(BMs) with two layers, one of hidden and another of visible neurons, and no
intra-layer connections. This property makes working with RBMs simpler than with regular
BMs, and in particular the stochastic computation of the log-likelihood gradient may be
performed more efficiently by means of Gibbs sampling
\cite{bengio-DeepArchitectures-2009-bk}.

In 2002, the \textit{Contrastive Divergence} learning algorithm (CD)
was proposed as an efficient training method for product-of-expert
models, from which RBMs are a special case
\cite{hinton-Contrastive-Divergence-2002-NC}. It was observed that
using CD to train RBMs worked quite well in practice. This fact was
important for deep learning since some authors suggested that a
multi-layer deep neural network is better trained when each layer is
pre-trained separately as if it were a single RBM
\cite{hinton-salakhutdinov-Auto-Encoders-2006-Science,
bengio-et-al-ContinuousInputs-And-AutoEncoders-2007-NIPS,
larochelle-et-al-Strategies-Train-Deep-Networks-2009-JMLR}. Thus,
training RBMs with CD and stacking up RBMs seems to be a good way to
go when designing deep learning architectures.

However, the picture is not as nice as it looks. CD is not a flawless
training algorithm. Despite CD being an approximation of the true
log-likelihood gradient
\cite{bengio-delalleau-Justifying-ContrastiveDivergence-2009-NC}, it
is biased and it may not converge in some cases
\cite{carreira-hinton-Contrastive-Divergence-Learning-2005-ICML,
  yuille-Convergence-Contrastive-Divergence-2005-NIPS,
  mackay-Failures-Contrastive-Divergence-2001-TR}. Moreover, it has
been observed that CD, and variants such as Persistent CD
\cite{tieleman-Persistent-Contrastive-Divergence-2008-ICML} or Fast
Persistent CD
\cite{tieleman-hinton-Improved-Persistent-Contrastive-Divergence-2009-ICML}
can lead to a steady decrease of the log-likelihood during learning
\cite{fischer-igel-Divergence-Contrastive-Divergence-2010-ICANN,
  desjardints-et-al-Parallel-Tempering-2010-AISTATS}. Therefore, the
risk of learning divergence imposes the requirement of a stopping
criterion. The two main methods used to decide when to stop the
learning process are \textit{reconstruction error} and
\textit{Annealed Importance Sampling} (AIS)
\cite{neal-Annealed-Importance-Sampling-1998-TR}. Reconstruction error
is easy to compute and it has been often used in practice, though its
adequacy remains unclear
\cite{fischer-igel-Divergence-Contrastive-Divergence-2010-ICANN}.  AIS
seems to work better than reconstruction error in some cases, though
it my also fail
\cite{schulz-et-al-Convergence-Contrastive-Divergence-2010-NIPSw}.

In this paper we propose an alternative stopping criteria for CD and
show its preliminary results.  These criteria are based on the
computation of two probabilities that do not require from the
knowledge of the partition function of the system. The early detection
of the decrease of the likelihood allows to overcome the
reconstruction error faulty observed behavior.

\section{Learning in Restricted Boltzmann Machines}
\label{learning}

\subsection{Energy-based Probabilistic Models}

Energy-based probabilistic models define a probability distribution
from an energy function, as follows:
\begin{equation}
\label{pdf-energy-xh}
 P(\x,\h) = \frac{\e^{-\text{Energy}(\x,\h)}}{Z} \ ,
\end{equation}
where $\x$ stand for visible variables and $\h$ are hidden variables
(typically binary) introduced to increase the expressive power of the
model. The normalization factor $Z$ is called partition function and reads
\begin{equation}
 Z = \sum_{\x,\h} \e^{-\text{Energy}(\x,\h)} \ .
\end{equation}

Since only $\x$ is observed, one is only interested in the
marginal distribution
\begin{equation}
\label{pdf-energy-x-sumh}
 P(\x) = \frac{\sum_{\h} \e^{-\text{Energy}(\x,\h)}}{Z} \ ,
\end{equation}
but the evaluation of the partition function $Z$ is
computationally prohibitive since it involves an exponentially large number
of terms.

The energy function depends on several parameters $\theta$, that are adjusted
at the learning stage. This is done by maximizing the
likelihood of the data. In energy-based models, the derivative of the
log-likelihood can be expressed as
\begin{eqnarray}
\label{dlog-likelihood}
\lefteqn{-\frac{\partial\log P(\x;\theta)}{\partial\theta} =  \ E_{P(\h|\x)} \left[\frac{ \partial\text{Energy}(\x,\h)}{\partial\theta}\right]} \nonumber  \\
{} & \ \ \ \ \ \ \ \
-\ E_{P(\xx)} \left[E_{P(\h|\xx)}\left[\frac{\partial\text{Energy}(\xx,\h)}{\partial\theta}\right] \right] \ ,
\end{eqnarray}
where the first term is called the positive phase and the second term is called the negative phase.
Similar to (\ref{pdf-energy-x-sumh}), the exact computation of the
derivative of the log-likelihood is usually unfeasible because of the
second term in (\ref{dlog-likelihood}), which comes from the
derivative of the partition function.

\subsection{Restricted Boltzmann Machines}

Restricted Boltzmann Machines are
energy-based probabilistic models whose energy function is:
\begin{equation}
\label{energy-RBM-discrete-binary-binary}
 \text{Energy}(\x,\h) = -\b^t\x - \c^t\h - \h^t\W\x \ .
\end{equation}

RBMs are at the core of DBNs
\cite{hinton-et-al-DeepBeliefNetworks-2006-NC} and other deep
architectures that use RBMs to unsupervised pre-training previous to
the supervised step
\cite{hinton-salakhutdinov-Auto-Encoders-2006-Science,bengio-et-al-ContinuousInputs-And-AutoEncoders-2007-NIPS,larochelle-et-al-Strategies-Train-Deep-Networks-2009-JMLR}.

The consequence of the particular form of the energy function is that
in RBMs both $P(\h|\x)$ and $P(\x|\h)$ factorize. In this way it is
possible to compute $P(\h|\x)$ and $P(\x|\h)$ in one step, making
possible to perform Gibbs sampling efficiently
\cite{geman-geman-Gibbs-Sampling-1984-TPAMI} that can be the basis of
the computation of an approximation of the derivative of the
log-likelihood (\ref{dlog-likelihood}).

\subsection{Contrastive Divergence}
\label{cd}
The most common learning algorithm for RBMs uses an algorithm to
estimate the derivative of the log-likelihood of a Product of
Experts model called CD \cite{hinton-Contrastive-Divergence-2002-NC}.

The algoritmh for CD$_n$ estimates the derivative of the log-likelihood as
\begin{eqnarray}
\label{dlog-likelihood-CD-n}
\lefteqn{-\frac{\partial\log P(\x_1;\theta)}{\partial\theta} \simeq \ E_{P(\h|\x_1)} \left[\frac{ \partial\text{Energy}(\x_1,\h)}{\partial\theta}\right]} \nonumber \\
{} & \ \ \ \ \ \ \ \
-\ E_{P(\h|\x_{n+1})}\left[\frac{\partial\text{Energy}(\x_{n+1},\h)}{\partial\theta}\right] \ .
\end{eqnarray}
where $\x_{n+1}$ is the last sample from the Gibbs chain starting
from $\x_1$ obtained after $n$ steps:
\begin{itemize}
\item[] $\h_1 \sim P(\h|\x_1)$
\item[] $\x_2 \sim P(\x|\h_1)$
\item[] ...
\item[] $\h_n \sim P(\h|\x_n)$
\item[] $\x_{n+1} \sim P(\x|\h_n)$ \ .
\end{itemize}

For binary RBMs,
$E_{P(\h|\x)}\left[\frac{\partial\text{Energy}(\x,\h)}{\partial\theta}\right]$
can be easily computed.

Several alternatives to CD$_n$ are Persistent CD (PCD)
\cite{tieleman-Persistent-Contrastive-Divergence-2008-ICML}, Fast PCD
(FPCD)
\cite{tieleman-hinton-Improved-Persistent-Contrastive-Divergence-2009-ICML}
or Parallel Tempering (PT)
\cite{desjardints-et-al-Parallel-Tempering-2010-AISTATS}.

\subsection{Monitoring the Learning Process in RBMs}

Learning in RBMs is a delicate procedure involving a lot of data
processing that one seeks to perform at a reasonable fast speed in
order to be able to handle large spaces with a huge amount of
states. In doing so, drastic approximations that can only be
understood in a statistically averaged sense are performed
(section~\ref{cd}).

One of the most relevant points to consider at the learning stage is
to find a good way to determine whether a good solution has been found
or not, and so to determine when should the learning process stop. One
of the most widely used criteria for stopping is the reconstruction
error, which is a measure of the capability of the network to produce
an output that is consistent with the data at input. Since RBMs are
probabilistic models, the reconstruction error of a data point
$\x^{(i)}$ is computed as the probability of $\x^{(i)}$ given the
expected value of $\h$ for $\x^{(i)}$:
\begin{equation}
\label{reconstruction-error-RBM-probability}
 R(\x^{(i)}) = P\left(\x^{(i)} | E\left[\h|\x^{(i)}\right]\right) \ ,
\end{equation}
which is the equivalent of the sum-of-squares reconstruction error for
deterministic networks.

Some authors have shown that it may happen that learning induces an
undesirable decrease in likelihood that goes undetected by the
reconstruction error
\cite{schulz-et-al-Convergence-Contrastive-Divergence-2010-NIPSw,
  fischer-igel-Divergence-Contrastive-Divergence-2010-ICANN}. It has
been studied
\cite{fischer-igel-Divergence-Contrastive-Divergence-2010-ICANN} that
the reconstruction error defined in
(\ref{reconstruction-error-RBM-probability}) usually decreases
monotonically. Since no increase in the reconstruction error takes
place during training there is no apparent way to detect the change of
behavior of the log-likelihood for CD$_n$.

\section{Proposed Stopping Criteria}
\label{proposed-stopping-criteria}

The proposed stopping criteria are based on the monitorization of the ratio of
two probabilities: the probability of the data (that should be high)
and the probability of points in the input space whose probability
should be low. More formally, it can be defined as:
\begin{equation}
\label{PXY_ek_prob_b1}
\xi = \frac{P(X)}{P(Y)} = \prod_{i=1}^N \frac{P(\x^{(i)})}{P(\y^{(i)})} \ ,
\end{equation}
where $X$ stands for the complete training set of $N$ samples and $Y$ is a suitable
artificially generated data set. The data set $Y$ can be generated in
different ways (see below).

The idea behind $\xi$ comes from the fact that the standard gradient
descent update rule used during learning requires from the evaluation
of two terms: the {\em positive} and {\em negative} phases. The
positive phase tends to decrease the energy (hence increase the
probability) of the states related to the training data, while the
negative phase tends to increase the energy of the whole set of
states with the corresponding decrease in probability. 
In this way, if $Y$ is selected so as to have low probability,
the numerator in $\xi$ is expected to increase while the denominator
is expected to decrease during the learning process, 
making $\xi$ maximal when learning is achieved.

Most relevant to the discussion is the fact that, being a ratio of
probabilities computed at every step of the Markov chain built
on-the-fly, the partition functions $Z$ involved in $P(X)$ and $P(Y)$
cancel out in $\xi$. In other words, the computation of $\xi$ can be
equivalently defined as
\begin{equation}
\label{PXY_ek_prob_b1_efficient}
\xi = \frac{P(X)}{P(Y)} =
  \prod_{i=1}^N \frac{\sum_{\h}\e^{-\text{Energy}(\x^{(i)},\h)}}
       {\sum_{\h}\e^{-\text{Energy}(\y^{(i)},\h)}} \ .
\end{equation}
The particular topology of RBMs allows to compute
$\sum_{\h}\e^{-\text{Energy}(\x,\h)}$ efficiently. This fact
dramatically decreases the computational cost involved in the
calculation, which would otherwise become unfeasible in most
real-world problems where RBMs could been successfully applied.

While $P(\x^{(i)})$ in $\xi$ is directly evaluated from the data in
the training set, the problem of finding suitable values for $Y$ still
remains. In order to select a point $\y^{(i)}$ with low probability,
one may seek for zones of the space distant from $\x^{(i)}$, thus
representing the complementary of the features to be learnt. This
point should not be difficult to find. On the one hand, in small
spaces one can enumerate the states. On the other hand, in large
spaces with a small training set $X$ the probability that a state picked up at
random does not belong to $X$ should be large. A second possibility is,
for fixed $\x^{(i)}$,
to suitably change 
the values of the hidden units 
during learning 
in such a way that they differ
significantly from the values they should take during data
reconstruction. 
We expect that, once learning is done, the reconstruction vectors
should be independent of the value of the hidden units. However, this
may not be the case while the system is still learning, as the basins 
of attraction of the energy functional depend explicitly on the values of the
weights and bias terms, which can change significantly.
This is in fact the main idea behind the stopping criteria proposed
in this work, that we shall exploit in the following.

With all that in mind, two different alternatives have been explored:
\begin{itemize}
\item[i)] $\y^{(i)} = E[\x|\h_s]$, where $\h_s$ is a random vector
  whose components are drawn from the uniform distribution in [0,1].
\item [ii)] $\y^{(i)} = E[\x|\h_s]$, where $\h_s$ = $1-\h^{(i)}_1$,
  i.e., the complementary of the first hidden vector obtained in the
  Gibbs chain for $\x^{(i)}$.
\end{itemize}

Regarding the first alternative, random hidden vectors are expected to
lead to regions of low reconstruction probability, at least while the
system is still learning. In the second alternative, we expect that if
a good reconstruction of $\x^{(i)}$ is achieved for a certain value of
$\h^{(i)}_1$ (see Eq.~(\ref{reconstruction-error-RBM-probability})),
the opposite should happen when $1-\h^{(i)}_1$ is used instead.


Other related possibilities like monitoring the average value
$E[\h|x^{(i)}_1]$ and using its complementary instead of $1-h^{(i)}_1$
have also been explored and yield similar results to the ones shown in
the following.

Notice that the reconstruction error only gathers information from the
training set $X$, while the proposed estimator $\xi$ in equation
(\ref{PXY_ek_prob_b1}) samples also states from the rest of the input
space.

\section{Experiments}
\label{experiments}

We performed several experiments to explore the aforementioned
alternatives defined in section~\ref{proposed-stopping-criteria}
and compare the behavior of the estimator $\xi$ to that of the actual
{\em log-likelihood} and the reconstruction error in a couple of
problems.

The first problem, denoted {\em Bars and Stripes} (BS), tries to
identify vertical and horizontal lines in 4$\times$4 pixel images. The
training set consists in the whole set of images containing all
possible horizontal or vertical lines (but not both), ranging from no
lines (blank image) to completely filled images (black image), thus
producing $2\times 2^4-2=30$ different images (avoiding the repetition
of fully back and fully white images) out of the space of $2^{16}$
possible images with black or white pixels.  The second problem, named
{\em Labeled Shifter Ensemble} (LSE), consists in learning 19-bit
states formed as follows: given an initial 8-bit pattern, generate
three new states concatenating to it the bit sequences 001, 010 or
100. The final 8-bit pattern of the state is the original one shifting
one bit to the left if the intermediate code is 001, copying it
unchanged if the code is 010, or shifting it one bit to the right if
the code is 100. One thus generates the training set using all
possible $2^8\times 3 = 768$ states that can be created in this form,
while the system space consists of all possible $2^{19}$ different
states one can build with 19 bits.  These two problems have already
been explored in
\cite{fischer-igel-Divergence-Contrastive-Divergence-2010-ICANN} and
are adequate in the current context since, while still large, the
dimensionality of space allows for a direct monitorization of the
partition function and the log-likelihood during learning.

In the following we discuss the learning processes of both problems
with single RBMs, employing the Contrastive Divergence algorithm
CD$_n$ with $n=1$ and $n=10$ as described in section~\ref{cd}. In all
cases, binary visible and hidden units were used.  In the BS case the
RBM had 16 visible and 8 hidden units, while in the LSE problem these
numbers were 19 and 10, respectively.  Every simulation was
carried out for a total of 50000 epochs, with measures being taken
every 50 epochs. Moreover, every point in the subsequent plots was the
average of ten different simulations starting from different random
values of the weights and bias. Other parameters affecting the results
that were changed along the analysis are the learning rate (LR)
involved in the weight and bias update rules and a weight decay
parameter (WD) that prevents weights from achieving large values that
would saturate the sigmoid functions present in the analytical
expressions associated to binary RBMs.

We present the results for the two problems at hand, showing for each
instance analyzed three different plots corresponding to the actual
log-likelihood of the problem, $log(\xi)$ ($\xi$ as defined in
(\ref{PXY_ek_prob_b1_efficient})) and the logarithm of the reconstructed
error (\ref{reconstruction-error-RBM-probability}), all three
quantities monitored during the learning process.

\begin{figure}[t!]
\begin{center}
\begin{tabular}{cc}
\hspace*{-0.8cm}
\includegraphics[width=0.30\textwidth]
  {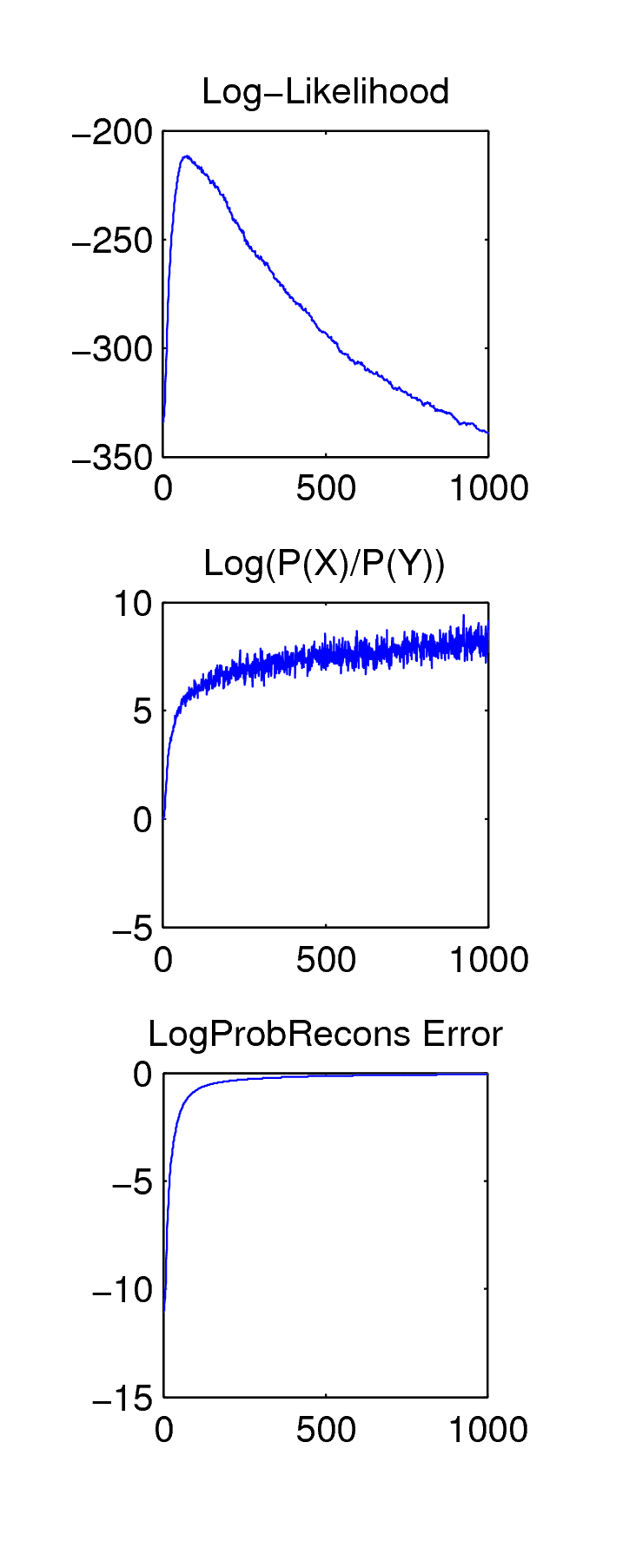} &
\hspace*{-0.8cm}
\hspace*{-0.5cm}
\includegraphics[width=0.30\textwidth]
  {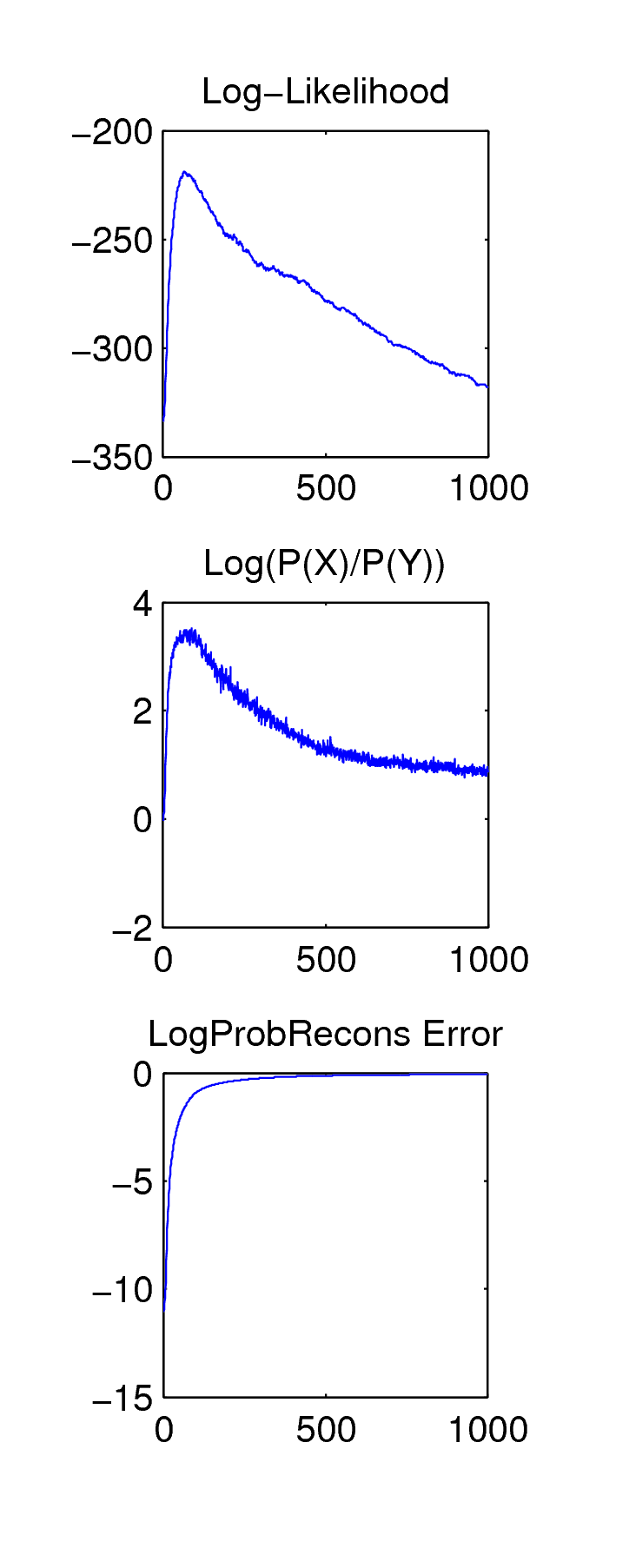}
\end{tabular}
\end{center}
\caption{
Log-likelihood, $\log(\xi)$ and log-probability of the reconstruction (upper, middle and lower 
panels) for the BS problem. Left and right columns correspond to options i) and ii)
when choosing values for the hidden units using CD$_1$ with LR=0.01 and WD=0. }
\label{fig_BS_1}
\end{figure}

\begin{figure}[t!]
\begin{center}
\begin{tabular}{cc}
\hspace*{-0.8cm}
\includegraphics[width=0.30\textwidth]
  {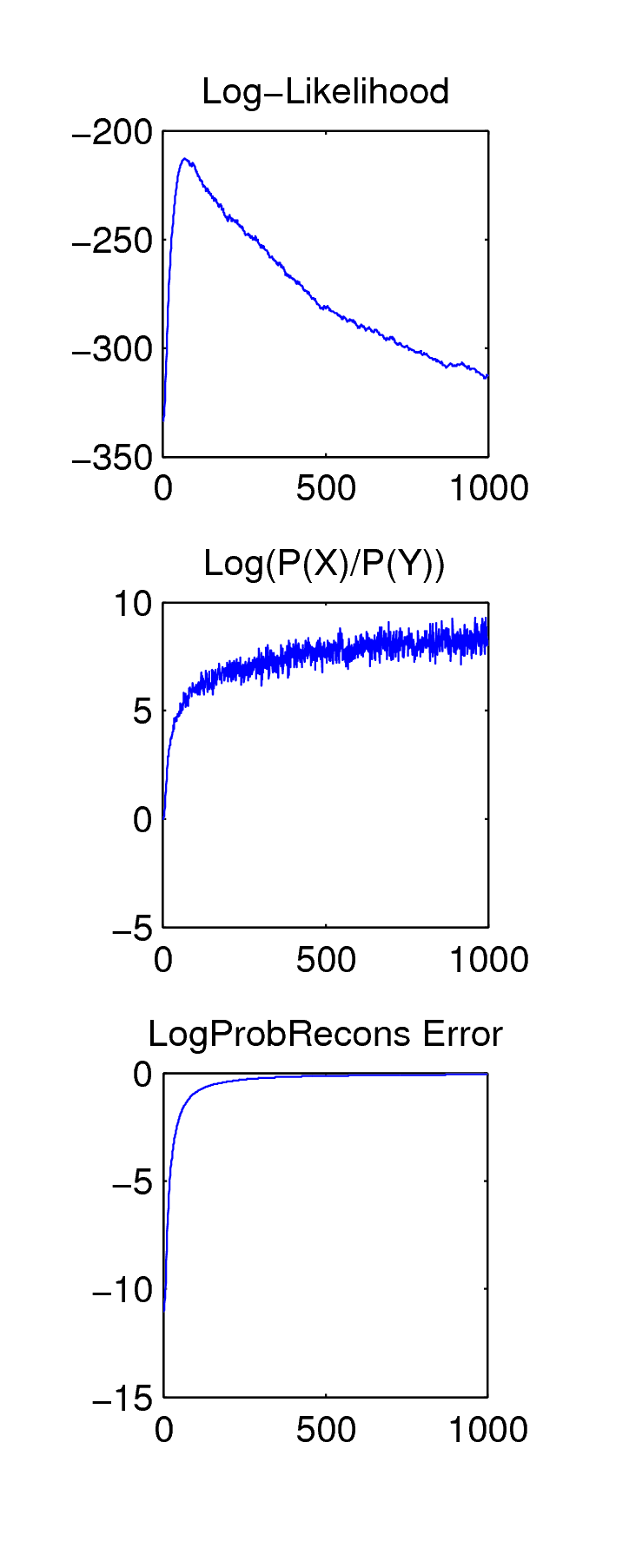} &
\hspace*{-0.8cm}
\hspace*{-0.5cm}
\includegraphics[width=0.30\textwidth]
  {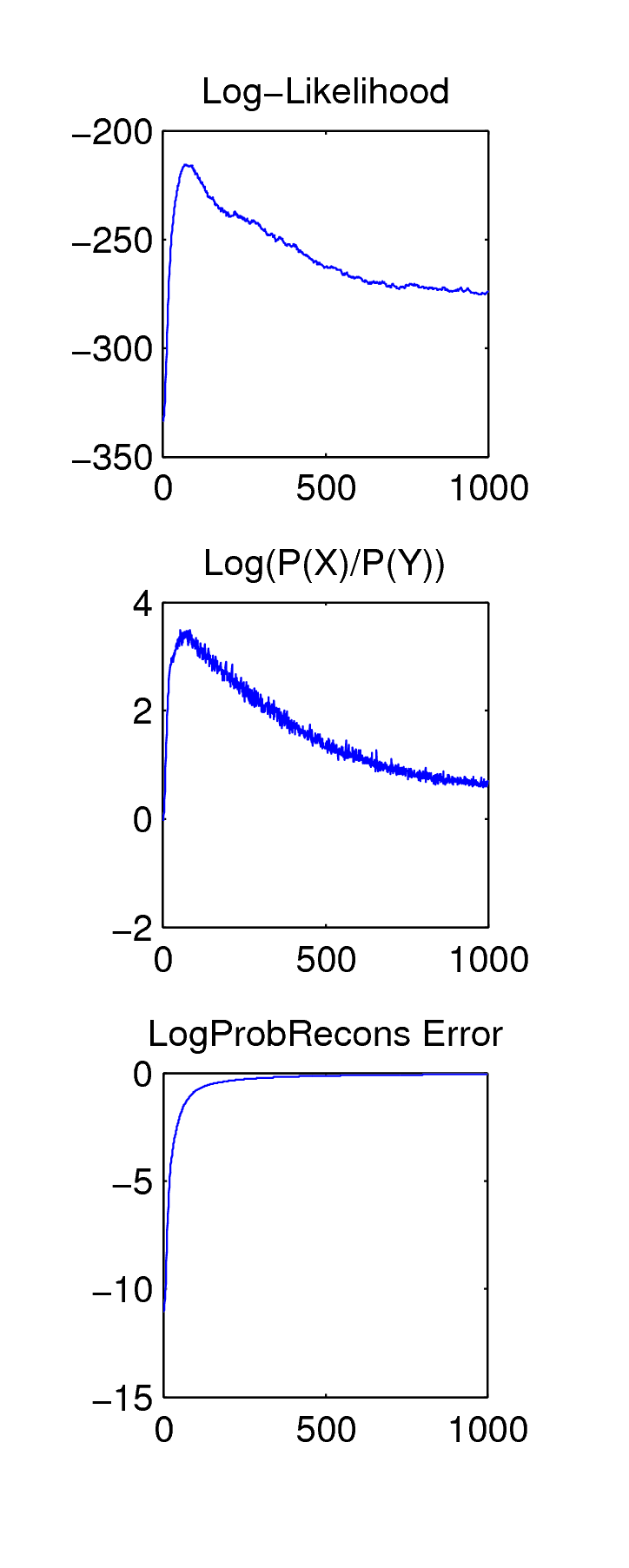}
\end{tabular}
\end{center}
\caption{
Same as in figure~\ref{fig_BS_1} but with WD=0.001.}
\label{fig_BS_2}
\end{figure}

Figure~\ref{fig_BS_1} shows results for the BS problem using the
alternatives i) and ii) defined in
section~\ref{proposed-stopping-criteria} using CD$_1$ with
LR=0.01 and WD=0. The left panel corresponds to alternative i) and the right
panel corresponds to alternative ii). As can be seen, the
log-likelihood increases very rapidly, reaches a maximum and then
starts decreasing, thus indicating that further learning only worsens
the model. In both cases, though, the log probability of the
reconstruction converges towards a constant value (very near 0,
indicating high probabilities for the reconstructed data), giving the
false impression that going on with the learning process will neither
improve nor worsen the predictions produced by the
network. Interestingly enough, though, the middle plot on the right
panel indicate that ii) is able to capture the increasing and
decreasing behavior of the log-likelihood, a feature that i) seems to
miss. At this point it looks like ii) is a better estimator of optimal
log-likelihood than the reconstruction error. This same behaviour is
seen in figure~\ref{fig_BS_2} where a weight decay value WD=0.001 is
employed.

\begin{figure}[t!]
\begin{center}
\begin{tabular}{cc}
\hspace*{-0.8cm}
\includegraphics[width=0.30\textwidth]
  {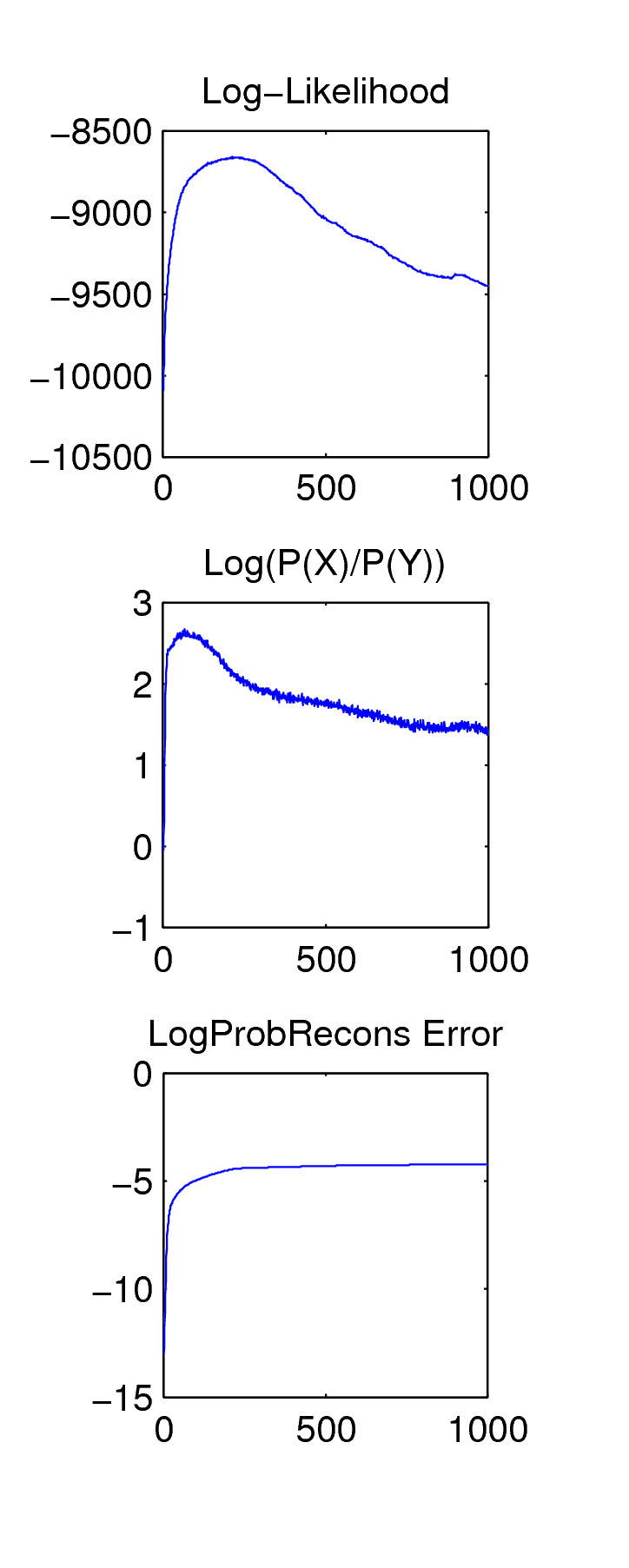} &
\hspace*{-0.8cm}
\hspace*{-0.5cm}
\includegraphics[width=0.30\textwidth]
  {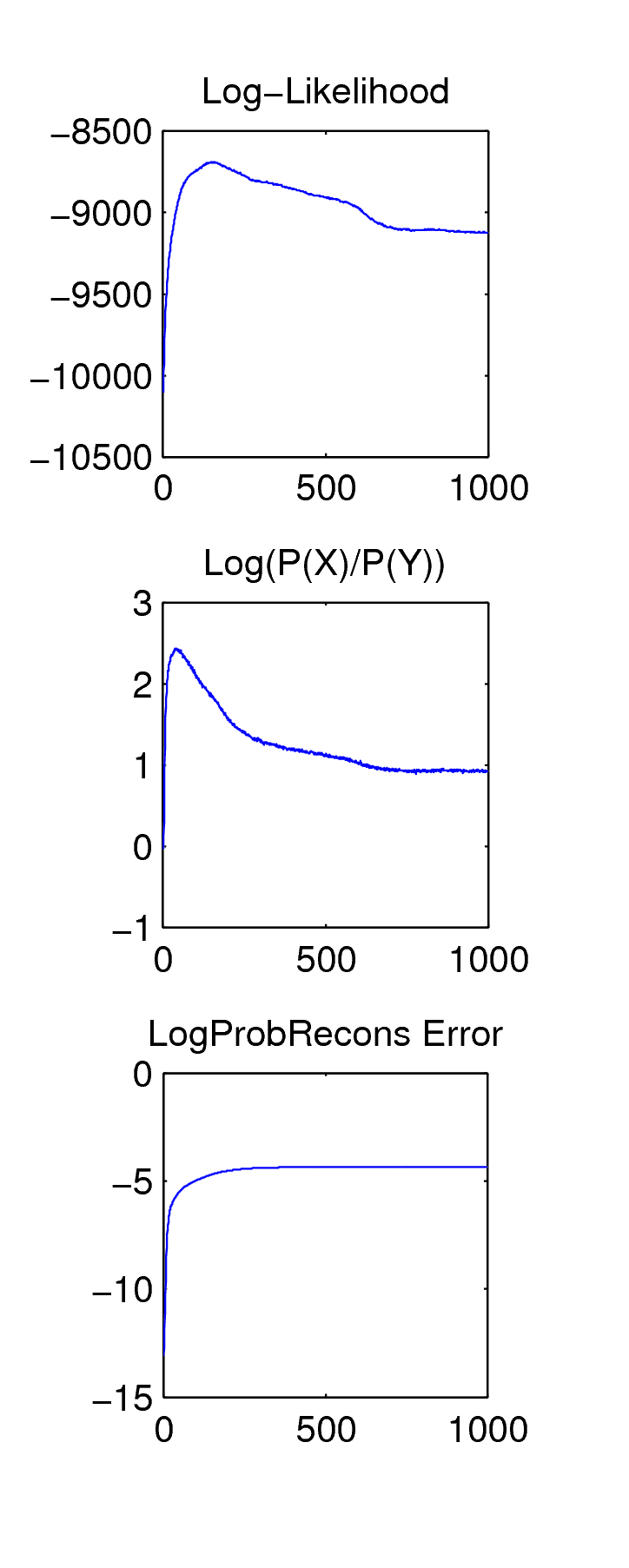}
\end{tabular}
\end{center}
\caption{
Results for the LSE problem as reported in figure~\ref{fig_BS_1} for CD$_1$, LR=0.001 and WD=0.001.}
\label{fig_LSE_1}
\end{figure}

The LSE problem yields somewhat similar results under the same
learning and monitoring conditions. The log-likelihood, $log(\xi)$ and
log-reconstruction error are shown as before in the upper, middle and
lower panels of figure~\ref{fig_LSE_1}, with options i) and ii) on the
left and right, respectively. In this case the learning rate has been
set to LR=0.001 (otherwise the log-likelihood of the problem decreases
monotonically). In this case, however, both estimators i) and ii) are
able to find the region where the log-likelihood is maximal,
decreasing similarly to the later when this point is surpassed.

These results seem to indicate that estimator ii) is more robust than
estimator i). Still, these two are better than the reconstruction
error which always present a similar pattern, both for the BS and LSE
problems, with a transient regime that always stabilizes to a plateau
that apparently has little to do with the actual behavior of the
log-likelihood.

\begin{figure}[t!]
\begin{center}
\begin{tabular}{cc}
\hspace*{-0.8cm}
\includegraphics[width=0.30\textwidth]
  {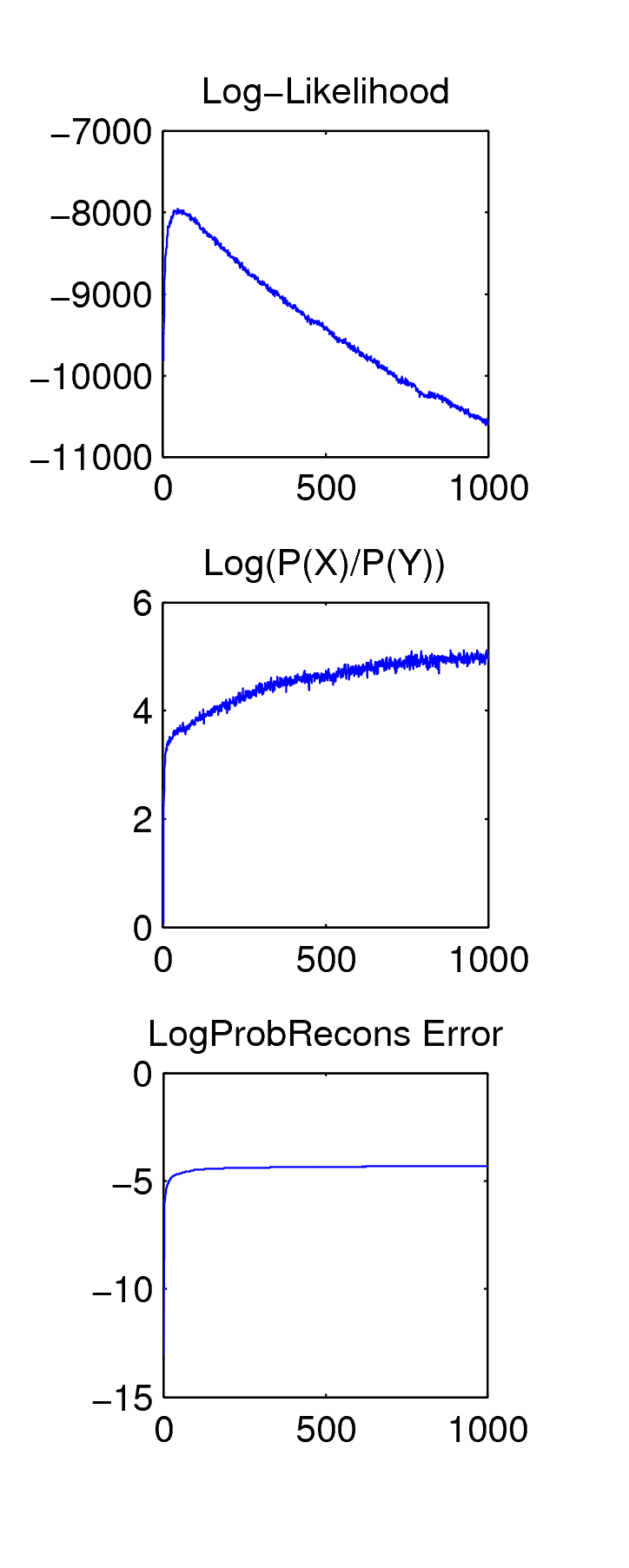} &
\hspace*{-0.8cm}
\hspace*{-0.5cm}
\includegraphics[width=0.30\textwidth]
  {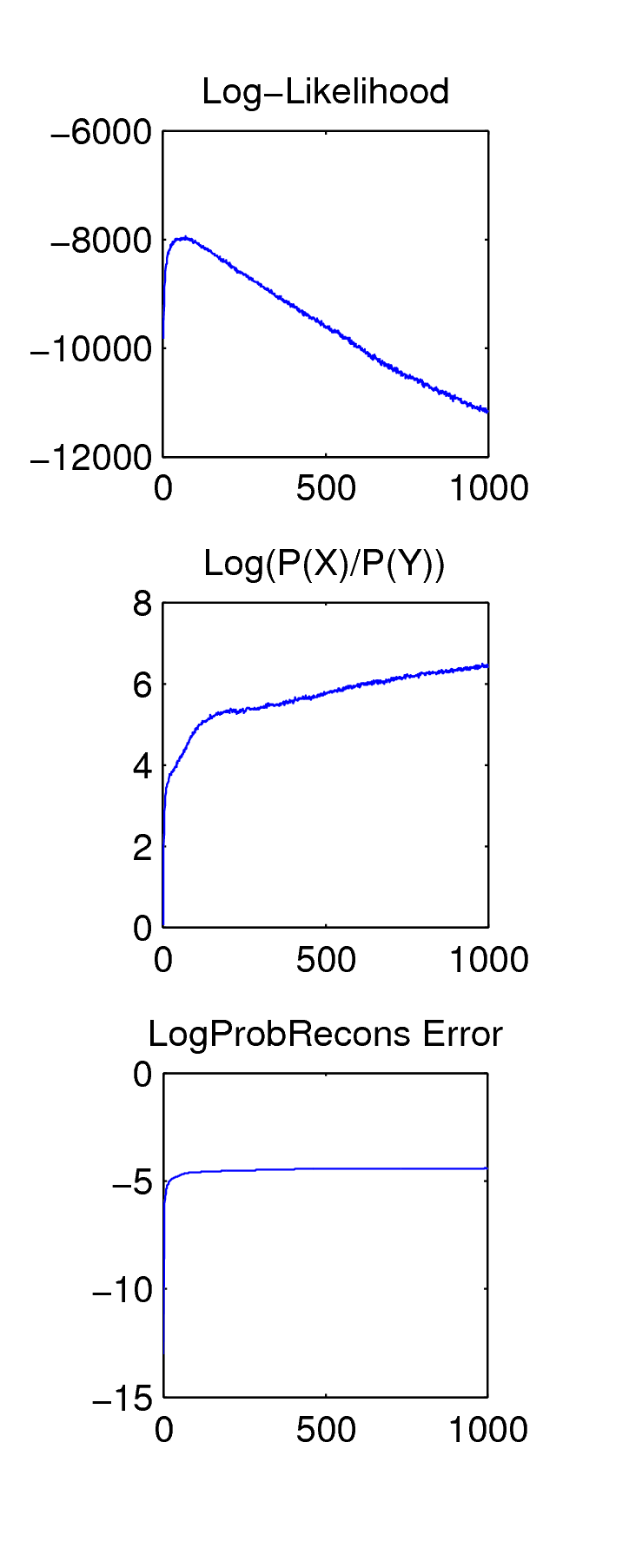}
\end{tabular}
\end{center}
\caption{
Results for the LSE problem as reported in figure~\ref{fig_BS_1} for CD$_{10}$, LR=0.01 and WD=0.}
\label{fig_LSE_2}
\end{figure}

All these results have been obtained in the CD$_1$
approximation. Since it is known that CD$_n$ with increasing $n$ can
lead to better learning results because of the increased statistical
independence of the input and output values generated, estimators i)
and ii) can also be used in this case. We have checked their
performance using CD$_{10}$ on the same two problems at hand. Results
for the LSE problem using CD$_{10}$, LR=0.01 and WD=0 are shown in
the left and right panels of figure~\ref{fig_LSE_2} for estimators i)
and ii), respectively. In this case, none of the estimators is able to
detect the region of maximal likelihood, stressing that none of these
shall be used as a test to stop the learning algorithm. However, the
reconstruction error has a similar behavior, thus indicating that it
is not a good testing quantity either. Similar results for the BS
problem are obtained when using CD$_{10}$. A possible explanation can
be related to the fact that the Markov chain involved in the process
tends to lose memory with increasing number of steps. Therefore, $\xi$
is computed with more independent data in CD$_{10}$ than in
CD$_{1}$. Anyway, the behavior of the proposed criteria with CD$_{10}$
should be further studied.

\begin{figure*}[t!]
\begin{center}
\hspace*{-0.8cm}
\includegraphics[width=1.0\textwidth]
  {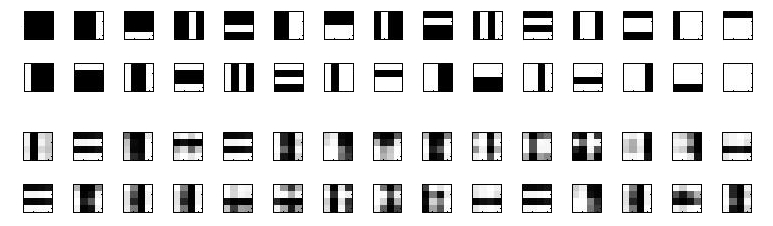}
\end{center}
\caption{ Training data (two upper rows) and generated samples (two
  lower rows) for the BS problems after 3000 epochs in the training
  process using CD$_1$.}
\label{barritas}
\end{figure*}

As a final remark, we note that for the BS problem the trained RBM
stopped using the proposed criteria is able to qualitatively generate
samples similar to those in the training set. We show in
figure~\ref{barritas} the complete training set (two upper rows) and
the same number of generated samples obtained from the RBM stopped
after 3000 epochs in the training process using CD$_1$ as discussed
above, corresponding to the maximum value of the proposed criterion
ii), which coincides with the optimal value of the log-likelihood (two
lower rows in the same figure).

\section{Conclusion}

Based on the fact that learning tries to increase the contribution of
the relevant states while decreasing the rest, two new estimators
based on the ratio of two probabilities have been proposed and
discussed as an alternative to the reconstruction error. It has been
shown that the better one, obtained by replacing the value of the
(binary) hidden units $\h$ by $1-\h$, can at some point be able to
monitor the actual behavior of the log-likelihood of the model without
additional computational cost. This estimator works well for CD$_1$
but for CD$_{10}$, which is considered to yield better learning
results at the expense however of a linear increase in computational
cost. We believe that the use of the estimator presented here in
CD$_1$ learning problems provides a faster stopping criteria for the
learning algorithm that can yield results compatible in quality to
those obtained in standard CD$_n$ learning for moderate $n$. Future
work along this line will be carried out in an attempt to formalize
that statement.

\section*{Acknowledgments} 
 
JD: This work was partially supported by MICINN project TIN2011-27479-C04-03 (BASMATI) and by SGR2009-1428 (LARCA).

FM: This work has been supported by grant No. FIS2011-25275 from DGI (Spain) and Grant No. 2009-SGR1003 from the Generalitat de Catalunya (Spain).

ER: This research is partially funded by Spanish research project TIN2012-31377.

\bibliography{StopCriteria}
\bibliographystyle{icml2014}

\end{document}